%% file: main.tex
\documentclass[10pt,twocolumn,letterpaper]{article}

\usepackage{cvpr}              

\input{preamble}

\definecolor{cvprblue}{rgb}{0.21,0.49,0.74}
\usepackage[pagebackref,breaklinks,colorlinks,citecolor=cvprblue]{hyperref}

\def\paperID{3929} 
\def\confName{CVPR}
\def\confYear{2024}

\title{En3D: An Enhanced Generative Model for Sculpting 3D Humans from \\
2D Synthetic Data}

\author{Yifang Men$^1$, Biwen Lei$^1$, Yuan Yao$^1$, Miaomiao Cui$^1$, Zhouhui Lian$^2$, Xuansong Xie$^1$ \\
{\tt\small $^1$Institute for Intelligent Computing, Alibaba Group} \\
{\tt\small $^2$Wangxuan Institute of Computer Technology, Peking University} \\
{\tt\small \url{https://menyifang.github.io/projects/En3D/index.html}} \\
}

\begin{document}

\twocolumn[{%
\renewcommand\twocolumn[1][]{#1}%
\maketitle

\vspace{-20 pt}
\begin{center}
    \centering
   \setlength{\abovecaptionskip}{0cm}
   \setlength{\belowcaptionskip}{0cm}
    \includegraphics[width=1.0\linewidth]{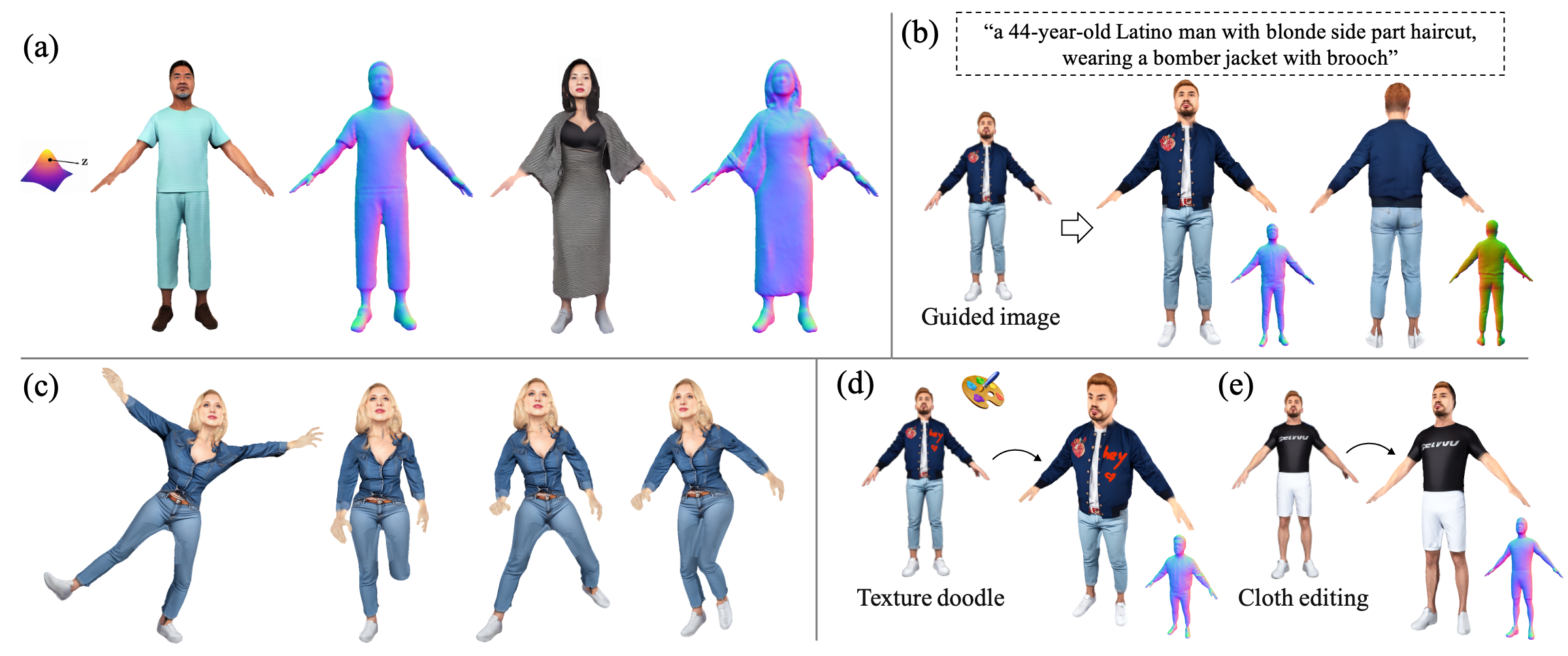}
    \captionof{figure}{Given random noises or guided texts, our generative scheme can synthesize high-fidelity 3D human avatars that are visually realistic and geometrically accurate. These avatars can be seamlessly animated and easily edited. Our model is trained on 2D synthetic data without relying on any pre-existing 3D or 2D collections.
}
    \label{fig:teaser}
\end{center}%
}]

\begin{abstract}
We present En3D, an enhanced generative scheme for sculpting high-quality 3D human avatars. Unlike previous works that rely on scarce 3D datasets or limited 2D collections with imbalanced viewing angles and imprecise pose priors, our approach aims to develop a zero-shot 3D generative scheme capable of producing visually realistic, geometrically accurate and content-wise diverse 3D humans without relying on pre-existing 3D or 2D assets. To address this challenge, we introduce a meticulously crafted workflow that implements accurate physical modeling to learn the enhanced 3D generative model from synthetic 2D data. During inference, we integrate optimization modules to bridge the gap between realistic appearances and coarse 3D shapes. Specifically, En3D comprises three modules: a 3D generator that accurately models generalizable 3D humans with realistic appearance from synthesized balanced, diverse, and structured human images; a geometry sculptor that enhances shape quality using multi-view normal constraints for intricate human anatomy; and a texturing module that disentangles explicit texture maps with fidelity and editability, leveraging semantical UV partitioning and a differentiable rasterizer. Experimental results show that our approach significantly outperforms prior works in terms of image quality, geometry accuracy and content diversity. We also showcase the applicability of our generated avatars for animation and editing, as well as the scalability of our approach for content-style free adaptation. 
\end{abstract}

\section{Introduction}
3D human avatars play an important role in various applications of AR/VR such as video games, telepresence and virtual try-on. 
Realistic human modeling is an essential task, and many valuable efforts have been made by leveraging neural implicit fields to learn high-quality articulated avatars 
~\cite{peng2021neural,dong2022pina,feng2022capturing,weng2022humannerf}.
However, these methods are directly learned from monocular videos or image sequences, where subjects are single individuals wearing specific garments, thus limiting their scalability. 

Generative models learn a shared 3D representation to synthesize clothed humans with varying identities, clothing and poses. Traditional methods are typically trained on 3D datasets, which are limited and expensive to acquire. This data scarcity limits the model’s generalization ability and may lead to overfitting on small datasets. Recently, 3D-aware image synthesis methods~\cite{or2022stylesdf,chan2022efficient,jo20233d} have demonstrated great potential in learning 3D generative models of rigid objects from 2D image collections. 
Follow-up works show the feasibility of learning articulated humans from image collections driven by SMPL-based deformations, but only in limited quality and resolution. EVA3D~\cite{hong2022eva3d} represents humans as a composition of multiple parts with NeRF representations. 
AG3D~\cite{dong2023ag3d} incorporates an efficient articulation module to capture both body shape and cloth deformation. Nevertheless, there remains a noticeable gap between generated and real humans in terms of appearance and geometry. Moreover, their results are limited to specific views (i.e., frontal angles) and lack diversity (i.e., fashion images in similar skin tone, body shape, and age). 

The aim of this paper is to propose a zero-shot 3D generative scheme that does not rely on any pre-existing 3D or 2D datasets, yet is capable of producing high-quality 3D humans that are visually realistic, geometrically accurate, and content-wise diverse. The generated avatars can be seamlessly animated and easily edited. An illustration is provided in Figure~\ref{fig:teaser}. 
To address this challenging task, our proposed method inherits from 3D-aware human image synthesis and exhibits substantial distinctions based on several key insights. 
Rethinking the nature of 3D-aware generative methods from 2D collections~\cite{chan2022efficient,hong2022eva3d,dong2023ag3d}, they actually try to learn a generalizable and deformable 3D representation, whose 2D projections can meet the distribution of human images in corresponding views. 
Thereby, it is crucial for accurate physical modeling between 3D objects and 2D projections. However, previous works typically leverage pre-existing 2D human images to estimate physical parameters (i.e., camera and body poses), which are inaccurate because of imprecise SMPL priors for highly-articulated humans. This inaccuracy limits the synthesis ability for realistic multi-view renderings. Second, these methods solely rely on discriminating 2D renderings, which is ambiguous and loose to capture inherent 3D shapes in detail, especially for intricate human anatomy.

To address these limitations, we propose a novel generative scheme with two core designs. 
Firstly, we introduce a meticulously-crafted workflow that implements accurate physical modeling to learn an enhanced 3D generative model from synthetic data. This is achieved by instantiating a 3D body scene and projecting the underlying 3D skeleton into 2D pose images using explicit camera parameters. These 2D pose images act as conditions to control a 2D diffusion model, synthesizing realistic human images from specific viewpoints. By leveraging synthetic view-balanced, diverse and structured human images, along with known physical parameters, we employ a 3D generator equipped with an enhanced renderer and discriminator to learn realistic appearance modeling.
Secondly, we improve the 3D shape quality by leveraging the gap between high-quality multi-view renderings and the coarse mesh produced by the 3D generative module. Specifically, we integrate an optimization module that utilizes multi-view normal constraints to 
rapidly refine geometry details under supervision.
Additionally, we incorporate an explicit texturing module to ensure faithful UV texture maps. In contrast to previous works that rely on inaccurate physical settings and inadequate shape supervision, we rebuild the generative scheme from the ground up, resulting in comprehensive improvements in image quality, geometry accuracy, and content diversity.
In summary, our contributions are threefold:
\begin{itemize}
\item We present a zero-shot generative scheme that efficiently synthesizes high-quality 3D human avatars with visual realism, geometric accuracy and content diversity. These avatars can be seamlessly animated and easily edited, offering greater flexibility in their applications.
\item We develop a meticulously-crafted workflow to learn an enhanced generative model from synthesized human images that are balanced, diverse, and also possess known physical parameters. This leads to diverse 3D-aware human image synthesis with realistic appearance.
\item We propose the integration of optimization modules into the 3D generator, leveraging multi-view guidance to enhance both shape quality and texture fidelity, thus achieving realistic 3D human assets.
\end{itemize}

\begin{figure*}
\begin{center}
\setlength{\abovecaptionskip}{0cm}
\setlength{\belowcaptionskip}{0cm}
\includegraphics[width=1\linewidth]{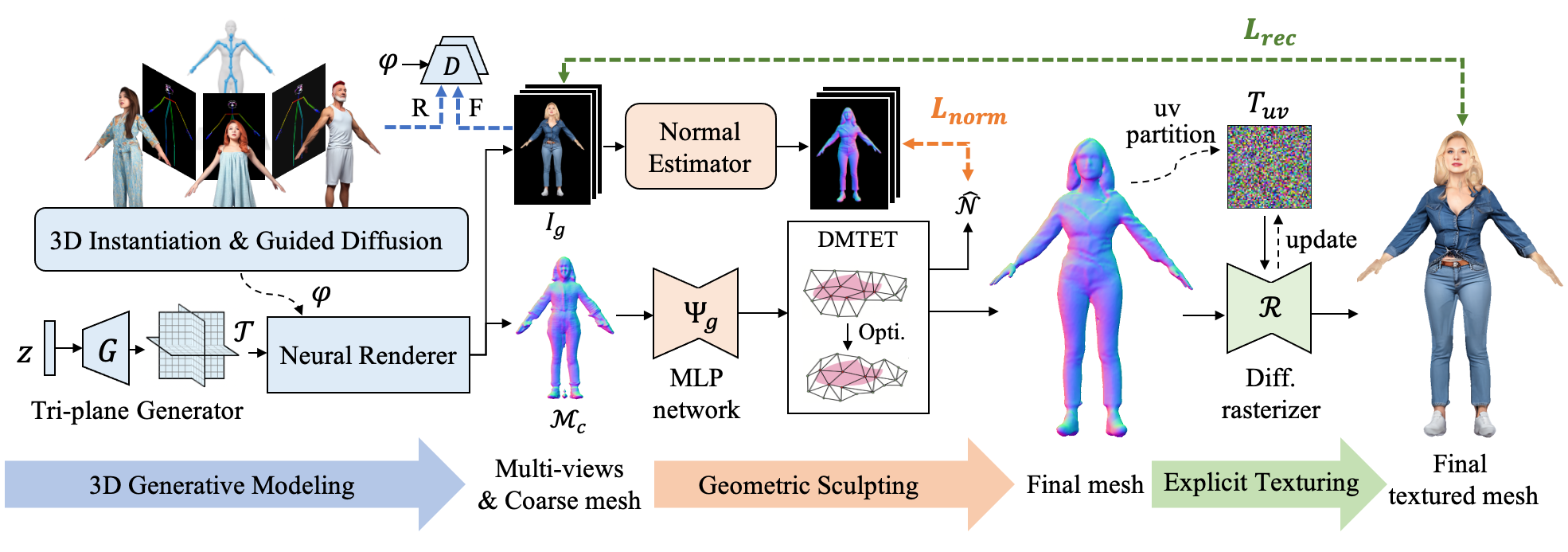}
\caption{An overview of the proposed scheme, which consists of three modules: 3D generative modeling (3DGM), the geometric sculpting (GS) and the explicit texturing (ET). 3DGM using synthesized diverse, balanced and structured human image with accurate camera $\varphi$ to learn generalizable 3D humans with the triplane-based architecture. GS is integrated as an optimization module by utilizing multi-view normal constraints to refine and carve geometry details. ET utilizes UV partitioning and a differentiable rasterizer to disentangles explicit UV texture maps. Not only multi-view renderings but also realistic 3D models can be acquired for final results.} 
\label{fig:network}
\end{center}
\end{figure*}

\section{Related work}

\noindent {\bf 3D Human Modeling.} Parametric models~\cite{anguelov2005scape,loper2015smpl,joo2018total,kanazawa2018end,osman2020star} serve as a common representation for 3D human modeling, they allows for robust control by deforming a template mesh with a series of low-dimensional parameters, but can only generate naked 3D humans. Similar ideas have been extended to model
clothed humans~\cite{alldieck2018video,ma2020learning}, but geometric expressivity is restricted due to the fixed mesh topology. 
Subsequent works~\cite{palafox2021npms,chen2022gdna,palafox2021npms} further introduce implicit surfaces to produce complex non-linear deformations of 3D bodies. Unfortunately, the aforementioned approaches all require 3D scans of various human poses for model fitting, which are difficult to acquire. 
With the explosion of NeRF, valuable efforts have been made towards combining NeRF models with explicit human models~\cite{peng2021neural,liu2021neural,dong2022pina,feng2022capturing,weng2022humannerf}. 
Neural body~\cite{peng2021neural} anchors a set of latent codes to the vertices of the SMPL model~\cite{loper2015smpl} and transforms the spatial locations of the codes to the volume in the observation space. 
HumanNeRF~\cite{weng2022humannerf} optimizes for a canonical, volumetric T-pose of the human with a motion field to map the non-rigid transformations. Nevertheless, these methods are learned directly from monocular videos or image sequences, where subjects are single individuals wearing specific garments, thus limiting their scalability.

\noindent {\bf Generative 3D-aware Image Synthesis.} 
Recently, 3D-aware image synthesis methods have lifted image generation with explicit view control by integrating the 2D generative models ~\cite{karras2019style,karras2020analyzing,karras2021alias} with 3D representations, such as voxels~\cite{wu2016learning,henzler2019escaping,nguyen2019hologan,nguyen2020blockgan}, meshes~\cite{szabo2019unsupervised,liao2020towards} and points clouds~\cite{li2019pu,achlioptas2018learning}. 
GRAF~\cite{schwarz2020graf} and $\pi$-GAN\cite{chan2021pi} firstly integrate the implicit representation networks, i.e., NeRF~\cite{mildenhall2021nerf}, with differentiable volumetric rendering for 3D scene generation.
However, they have difficulties in training on high-resolution images due to the costly rendering process. 
Subsequent works have sought to improve the efficiency and quality of such NeRF-based GANs, either by adopting a two-stage rendering process~\cite{gu2021stylenerf,niemeyer2021giraffe,or2022stylesdf,chan2022efficient,xue2022giraffe} or a smart sampling strategy~\cite{deng2022gram,zhou2021cips}. 
StyleSDF~\cite{or2022stylesdf} combines a SDF-based volume renderer and a 2D StyleGAN network~\cite{karras2020analyzing} for photorealistic image generation. 
EG3D~\cite{chan2022efficient} introduces a superior triplane representation to leverage 2D CNN-based feature generators for efficient generalization over 3D spaces. 
Although these methods demonstrate impressive quality in view-consistent image synthesis, they are limited to simplified rigid objects such as faces, cats and cars.

To learn highly articulated humans from unstructured 2D images, recent works~\cite{zhang2022avatargen,hong2022eva3d,fu2022stylegan,jiang2023humangen,dong2023ag3d,yang20233dhumangan} integrate the deformation field to learn non-rigid deformations based on the body prior of estimated SMPL parameters.  
EVA3D~\cite{hong2022eva3d} represents humans as a composition of multiple parts with NeRF representations. 
Instead of directly rendering the image from a 3D representation, 3DHumanGAN~\cite{yang20233dhumangan} uses an equivariant 2D generator modulated by 3D human body prior, which enables to establish one-to-many mapping from 3D geometry to synthesized textures from 2D images. 
AG3D~\cite{dong2023ag3d} combines the 3D generator with an efficient articulation module to warp from canonical space into posed space via a learned continuous deformation field. 
However, a gap still exists between the generated and real humans in terms of appearance, due to the imprecise priors from complex poses as well as the data biases from limited human poses and imbalanced viewing angles in the dataset.


\section{Method Description}

Our goal is to develop a zero-shot 3D generative scheme that does not rely on any pre-existing 3D or 2D collections, yet is capable of producing high-quality 3D humans that are visually realistic, geometrically accurate and content-wise diverse to generalize to arbitrary humans. 

An overview of the proposed scheme is illustrated in Figure~\ref{fig:network}. We build a sequential pipeline with the following three modules: the 3D generative modeling (3DGM), the geometric sculpting (GS) and the explicit texturing (ET). 
The first module synthesizes view-balanced, structured and diverse human images with known camera parameters. Subsequently, it learns a 3D generative model from these synthetic data, focusing on realistic appearance modeling (Section~\ref{sec:3dgm}). To overcome the inaccuracy of the 3D shape, the GS module is incorporated during the inference process. It optimizes a hybrid representation with multi-view normal constraints to carve intricate mesh details (Section~\ref{sec:gs}). Additionally, the ET module is employed to disentangle explicit texture by utilizing semantical UV partitioning and a differentiable rasterizer (Section~\ref{sec:et}). By combining these modules, we are able to synthesize high-quality and faithful 3D human avatars by incorporating random noises or guided texts/images (Section~\ref{sec:infer}).

\subsection{3D generative modeling}
\label{sec:3dgm}
Without any 3D or 2D collections, we develop a synthesis-based flow to learn a 3D generative module from 2D synthetic data. 
We start by instantiating a 3D scene through the projection of underlying 3D skeletons onto 2D pose images, utilizing accurate physical parameters (i.e., camera parameters). 
Subsequently, the projected 2D pose images serve as conditions to control the 2D diffusion model~\cite{zhang2023adding} for synthesizing view-balanced, diverse, and lifelike human images. 
Finally, we employ a triplane-based generator with enhanced designs to learn a generalizable 3D representation from the synthetic data.
Details are described as follows.

\noindent {\bf 3D instantiation.} 
Starting with a template body mesh (e.g., SMPL-X~\cite{pavlakos2019expressive}) positioned and posed in canonical space, we estimate the 3D joint locations $\mathcal{P}_{3d}$ by regressing them from interpolated vertices. We then project $\mathcal{P}_{3d}$ onto 2D poses $\mathcal{P}_i, i=1,…,k$ from $\mathcal{K}$ horizontally uniformly sampled viewpoints $\varphi$. In this way, paired 2D pose images and their corresponding camera parameters $\{\mathcal{P}_i, \varphi_i \}$ are formulated.

\noindent {\bf Controlled 2D image synthesis.}
With the pose image $\mathcal{P}_i$, we feed it into off-the-shelf ControlNet~\cite{zhang2023adding} as the pose condition to guide diffusion models ~\cite{rombach2022high} to synthesize human images in desired poses (i.e., views). 
The text prompt $T$ is also used for diverse contents. Given a prompt $T$, instead of generating a human image $\mathcal{I}_s: \mathcal{I}_s = \mathcal{C}(\mathcal{P}_i, T)$ independently for each view $\varphi_i$, we horizontally concatenate $\mathcal{K}$ pose images $\mathcal{P}_i \in R^{H\times W\times 3}$, resulting in $\mathcal{P}_i' \in R^{H\times KW\times 3}$ and feed $\mathcal{P}_i’$ to $\mathcal{C}$, along with a prompt hint of ‘multi-view’ in $T$. In this way, multi-view human images $\mathcal{I}_s'$ are synthesized with roughly coherent appearance. We split $\mathcal{I}_s'$ to single view images $\mathcal{I}_\varphi$ under specific views $\varphi$.
This concatenation strategy facilitates the convergence of distributions in synthetic multi-views, thus easing the learning of common 3D representation meeting multi-view characteristics.

\noindent {\bf Generalizable 3D representation learning.}
With synthetic data of paired $\{\mathcal{I}_\varphi, \varphi \}$, we learn the 3D generative module $\mathcal{G}_{3d}$ from them to produce diverse 3D-aware human images with realistic appearance. 
Inspired by EG3D~\cite{chan2022efficient}, 
we employ a triplane-based generator to produce a generalizable representation $\mathcal{T}$ and introduce a patch-composed neural renderer to learn intricate human representation efficiently.
Specifically, instead of uniformly sampling 2D pixels on the image $\mathcal{I}$, we decompose patches in the ROI region including human bodies, and only emit rays towards pixels in these patches.
The rays are rendered into RGB color with opacity values via volume rendering. 
Based on the decomposed rule, we decode rendered colors to multiple patches and re-combine these patches for full feature images. 
In this way, the representation is composed of effective human body parts, which directs the attention of the networks towards the human subject itself. 
This design facilitates fine-grained local human learning while maintaining computational efficiency.

For the training process, we employ two discriminators, one for RGB images and another for silhouettes, which yields better disentanglement of foreground objects with global geometry. 
The training loss for this module $L_{3d}$ consists of the two adversarial terms:
\begin{equation}
\mathcal{L}_{3d}= \mathcal{L}_{adv}(\mathcal{D}_{rgb}, \mathcal{G}_{3d})+ \lambda_{s} \mathcal{L}_{adv} (\mathcal{D}_{mask}, \mathcal{G}_{3d}),
\end{equation}
where $\lambda_s$ denotes the weight of silhouette item. $\mathcal{L}_{adv}$ is computed by the non-saturating GAN loss with R1 regularization~\cite{mescheder2018training}. 

With the trained $\mathcal{G}_{3d}$, we can synthesize 3D-aware human images $\mathcal{I}_g^\varphi$ with view control, and extract coarse 3D shapes $\mathcal{M}_c$ from the density field of neural renderer using the Marching Cubes algorithm~\cite{lorensen1998marching}. 

\subsection{Geometric sculpting}
\label{sec:gs}
Our 3D generative module can produce high-quality and 3D-consistent human images in view controls.
However, its training solely relies on discriminations made using 2D renderings, which can result in inaccuracies in capturing the inherent geometry, especially for complex human bodies. 
Therefore, we integrate the geometric sculpting, an optimization module leveraging geometric information from high-quality multi-views to carve surface details. Combined with a hybrid 3D representation and a differentiable rasterizer, it can rapidly enhance the shape quality within seconds.

\noindent {\bf DMTET adaption.} Owing to the expressive ability of arbitrary topologies and computational efficiency with direct shape optimization, we employ DMTET as our 3D representation in this module and adapt it to the coarse mesh $\mathcal{M}_{c}$ via an initial fitting procedure. Specifically, we parameterize DMTET as an MLP network $\Psi_g$ that learns to predict the SDF value $s(v_i)$ and the position offset $\delta v_i$ for each vertex $v_i \in VT$ of the tetrahedral grid $(VT , T )$. A point set $P = \{p_i\in R^3 \}$ is randomly sampled near $\mathcal{M}_{c}$ and their SDF values $SDF(p_i)$ can be pre-computed. We adapt the parameters $\psi$ of $\Psi_g$ by fitting it to the SDF of $\mathcal{M}_{c}$: 
\begin{equation}
\mathcal{L}_{ada} = \sum_{p_i\in P}||s(p_i; \psi)-SDF(p_i)||_2.
\end{equation}

\noindent {\bf Geometry refinement.} 
Using the adapted DMTET, we leverage the highly-detailed normal maps $\mathcal{N}$ derived from realistic multi-view images as a guidance to refine local surfaces. 
To obtain the pseudo-GT normals $\mathcal{N_\varphi}$, we extract them from $\mathcal{I}_g^\varphi$ using a pre-trained normal estimator~\cite{xiu2022icon}. 
For the rendered normals $\mathcal{\hat{N}_\varphi}$, we extract the triangular mesh $\mathcal{M}_{tri}$ from $(VT, T)$ using the Marching Tetrahedra (MT) layer in our current DMTET. By rendering the generated mesh $\mathcal{M}_{tri}$ with differentiable rasterization, we obtain the resulting normal map $\mathcal{\hat{N}_\varphi}$.
To ensure holistic surface polishing that takes into account multi-view normals, we randomly sample camera poses $\varphi$ that are uniformly distributed in space. We optimize the parameters of $\Psi_g$ using the normal loss, which is defined as:
\begin{equation}
\mathcal{L}_{norm} = ||\mathcal{\hat{N}_\varphi}-\mathcal{N_\varphi}||_2.
\end{equation}

After rapid optimization, the final triangular mesh $\mathcal{M}_{tri}$ can be easily extracted from the MT layer. 
If the hands exhibit noise, they can be optionally replaced with cleaner geometry hands from SMPL-X, benefiting from the alignment of the generated body in canonical space with the underlying template body.

\subsection{Explicit texturing}
\label{sec:et}

With the final mesh, the explicit texturing module aims to disentangle a UV texture map from multi-view renderings $\mathcal{I}_g^\varphi$.
This intuitive module not only facilitates the incorporation of high-fidelity textures but also enables various editing applications, as verified in Section~\ref{sec:application}.
 
Given the polished triangular mesh $\mathcal{M}_{tri}$ and multi-views ${\mathcal{I}_g^\varphi}$, 
we model the explicit texture map $T_{uv}$ of $\mathcal{M}_{tri}$ with a semantic UV partition and optimize $T_{uv}$ using a differentiable rasterizer $\mathcal{R}$~\cite{laine2020modular}.
Specifically, leveraging the canonical properties of synthesized bodies, we semantically split $\mathcal{M}_{tri}$ into $\gamma$ components and rotate each component vertically, thus enabling effective UV projection for each component with cylinder unwarping. 
We then combine the texture partitions together for the full texture $T_{uv}$.
We optimize $T_{uv}$ from a randomly initialized scratch using the texture loss, which consists of a multi-view reconstruction term and a total-variation (tv) term:
\begin{equation}
\mathcal{L}_{tex} = \mathcal{L}_{rec} + \lambda_{tv} \mathcal{L}_{tv},
\end{equation}
where $\lambda_{tv}$ denotes the weight of the tv loss.

\noindent {\bf Multi-view guidance.}
To ensure comprehensive texturing in the 3D space, we render the color images $\mathcal{R}(\mathcal{M}_{tri}, \varphi)$ and silhouettes $\mathcal{S}$ using $\mathcal{R}$ and optimize $T_{uv}$ utilizing multi-view weighted guidance. Their pixel-alignment distances to the original multi-view renderings $\mathcal{I}_g^\varphi$ are defined as the reconstruction loss:
\begin{equation}
\mathcal{L}_{rec} = \sum_{\varphi\in \Omega}w_\varphi||\mathcal{R}(\mathcal{M}_{tri}, \varphi)\cdot \mathcal{S}-I_g^\varphi\cdot \mathcal{S}||_2,
\end{equation}
where $\Omega$ is the set of viewpoints $\{\varphi_i, i=1,...,k\}$ and $w_\varphi$ denotes weights of different views. $w_\varphi$ equals to $1.0$ for $\varphi\in \{front, back\}$ and $0.2$ otherwise.

\noindent {\bf Smooth constraint.} To avoid abrupt variations and smooth the generated texture $T_{uv}$, we utilize the total-variation loss $\mathcal{L}_{tv}$ which is computed by: 
\begin{equation}
 \mathcal{L}_{tv} = \frac{1}{h\times w\times c}||\nabla_x(T_{uv})+\nabla_y(T_{uv})||,
\end{equation}
where $x$ and $y$ denote horizontal and vertical directions. 

\begin{figure}
\begin{center}
\setlength{\abovecaptionskip}{0.1cm}
\setlength{\belowcaptionskip}{-0.2cm}
\includegraphics[width=1\linewidth]{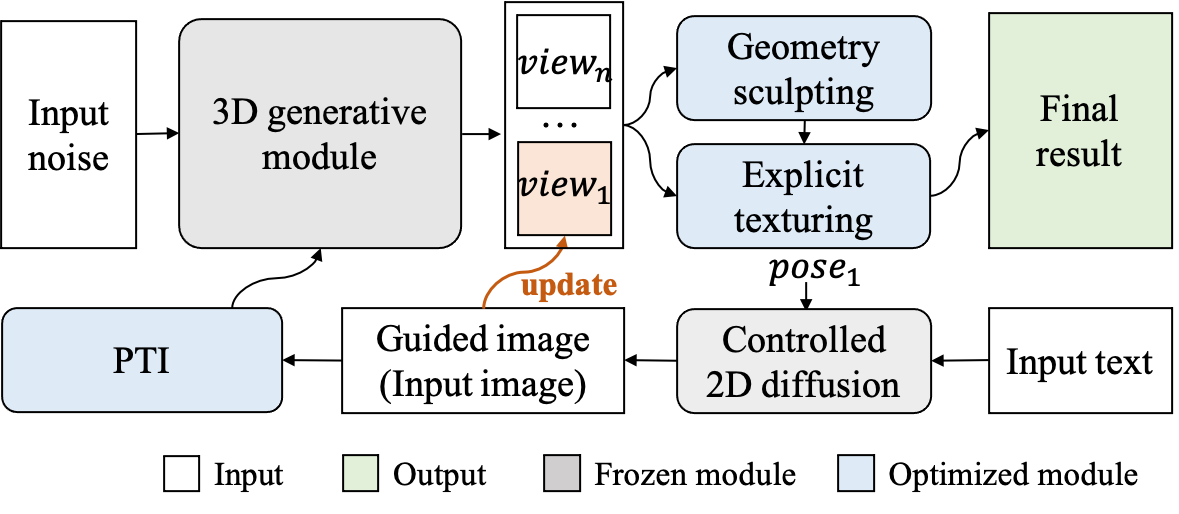}
\caption{The visualized flowchart of our method that synthesize textured 3D human avatars from input noises, texts or images.}
\label{fig:infer}
\end{center}
\end{figure}

\subsection{Inference}
\label{sec:infer}
Built upon the above modules, we can generate high-quality 3D human avatars from either random noises or guided inputs such as texts or images. 
The flowchart for this process is shown in Figure~\ref{fig:infer}.
For input noises, we can easily obtain the final results by sequentially using the 3DGM, GS and ET modules. For text-guided synthesis, we first convert the text into a structured image using our controlled diffusion $\mathcal{C}$, and then inverse it to the latent space using PTI~\cite{roich2022pivotal}. 
Specially, the GS and ET modules provide an interface that accurately reflects viewed modifications in the final 3D objects. As a result, we utilize the guided image to replace the corresponding view image, which results in improved fidelity in terms of geometry and texture. 
The same process is applied for input images as guided images.


\section{Experimental Results}

\begin{figure*}
\begin{center}
\setlength{\abovecaptionskip}{0.2cm}
\setlength{\belowcaptionskip}{-0.2cm}
\includegraphics[width=1\linewidth]{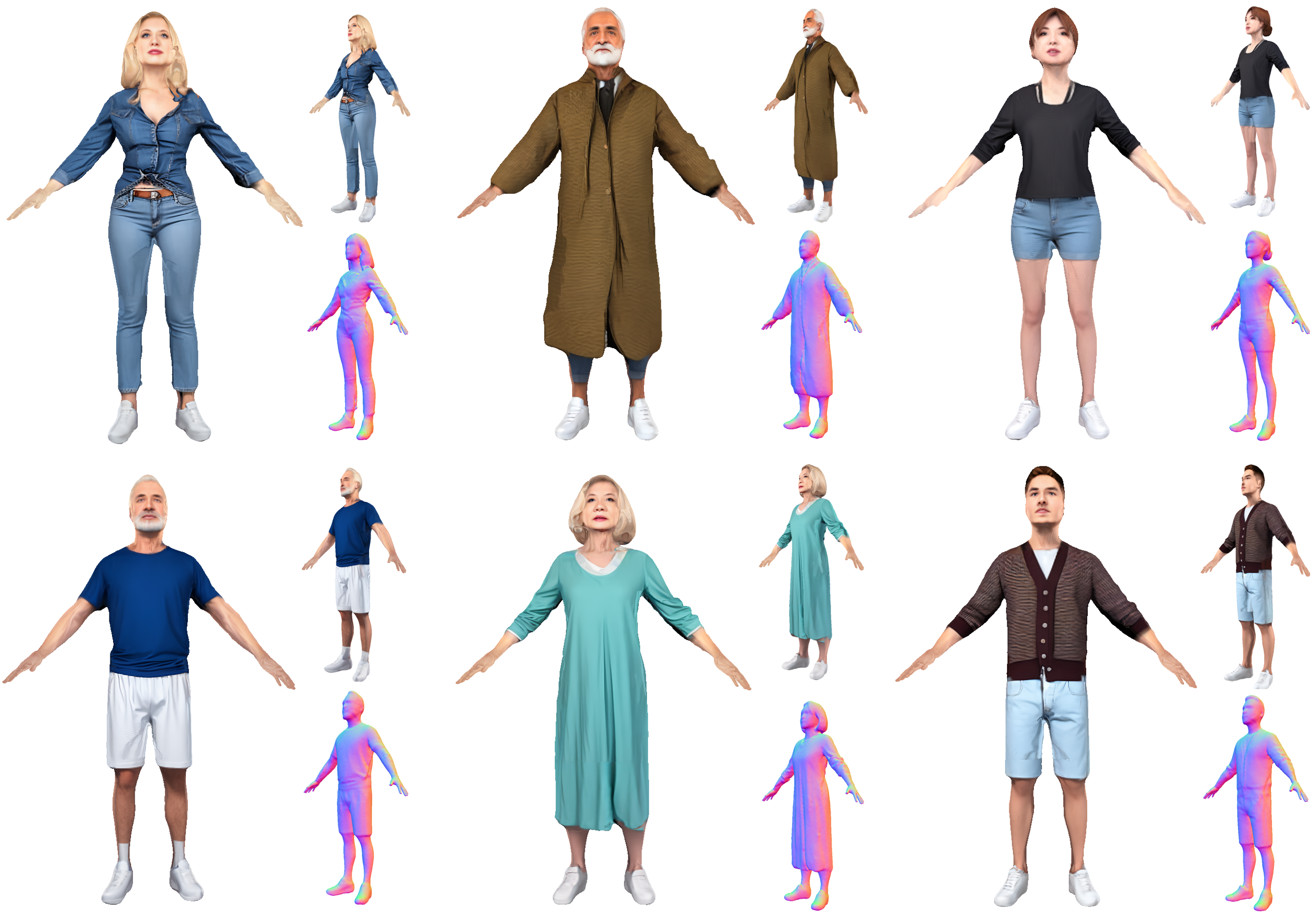}
\caption{Results of synthesized 3D human avatars at $512^2$. }
\label{fig:synthesis}
\end{center}
\end{figure*}

\noindent {\bf Implementation details.} 
Our process begins by training the 3D generative module (3DGM) on synthetic data. During inference, we integrate the geometric sculpting (GS) and explicit texturing (ET) as optimization modules. 
For 3DGM, we normalize the template body to the $(0,1)$ space and place its center at the origin of the world coordinate system. 
We sample $7 (\mathcal{K}=7)$ viewpoints uniformly from the horizontal plane, ranging from $0^{\circ}$ to $180^{\circ}$ (front to back), with a camera radius of $2.7$.
For each viewpoint, we generate $100K$ images using the corresponding pose image. 
To ensure diverse synthesis, we use detailed descriptions of age, gender, ethnicity, hairstyle, facial features, and clothing, leveraging a vast word bank. 
To cover $360^{\circ}$ views, we horizontally flip the synthesized images and obtain 1.4 million human images at a resolution of $512^2$ in total. 
We train the 3DGM for about 2.5M iterations with a batch size of $32$, using two discriminators with a learning rate of $0.002$ and a generator learning rate of 0.0025. The training takes 8 days on 8 NVIDIA Tesla-V100. 
For GS, we optimize $\psi$ for 400 iterations for DMTET adaption and 100 iterations for surface carving (taking about 15s in total on 1 NVIDIA RTX 3090 GPU). For ET, we set $\lambda_{uv}=1$ and optimize $T_{uv}$ for 500 iterations (around 10 seconds). 
We split $\mathcal{M}_{tri}$ into $5 (\gamma=5)$ body parts (i.e., trunk, left/right arm/leg) with cylinder UV unwarping. We use the Adam optimizer with learning rates of 0.01 and 0.001 for $\Psi_g$ and $T_{uv}$, respectively. 
Detailed network architectures can be found in the supplemental materials (Suppl).

\subsection{3D human generation}
Figure~\ref{fig:synthesis} showcases several 3D human avatars synthesized by our pipeline, highlighting the image quality, geometry accuracy, and diverse outputs achieved through our method. Additionally, we explore the interpolation of the latent conditions to yield smooth transitions in appearance, leveraging the smooth latent space learned by our generative model. 
For more synthesized examples and interpolation results, please refer to the Suppl.

\begin{figure*}
\begin{center}
\setlength{\abovecaptionskip}{0.1cm}
\setlength{\belowcaptionskip}{-0.2cm}
\includegraphics[width=1\linewidth]{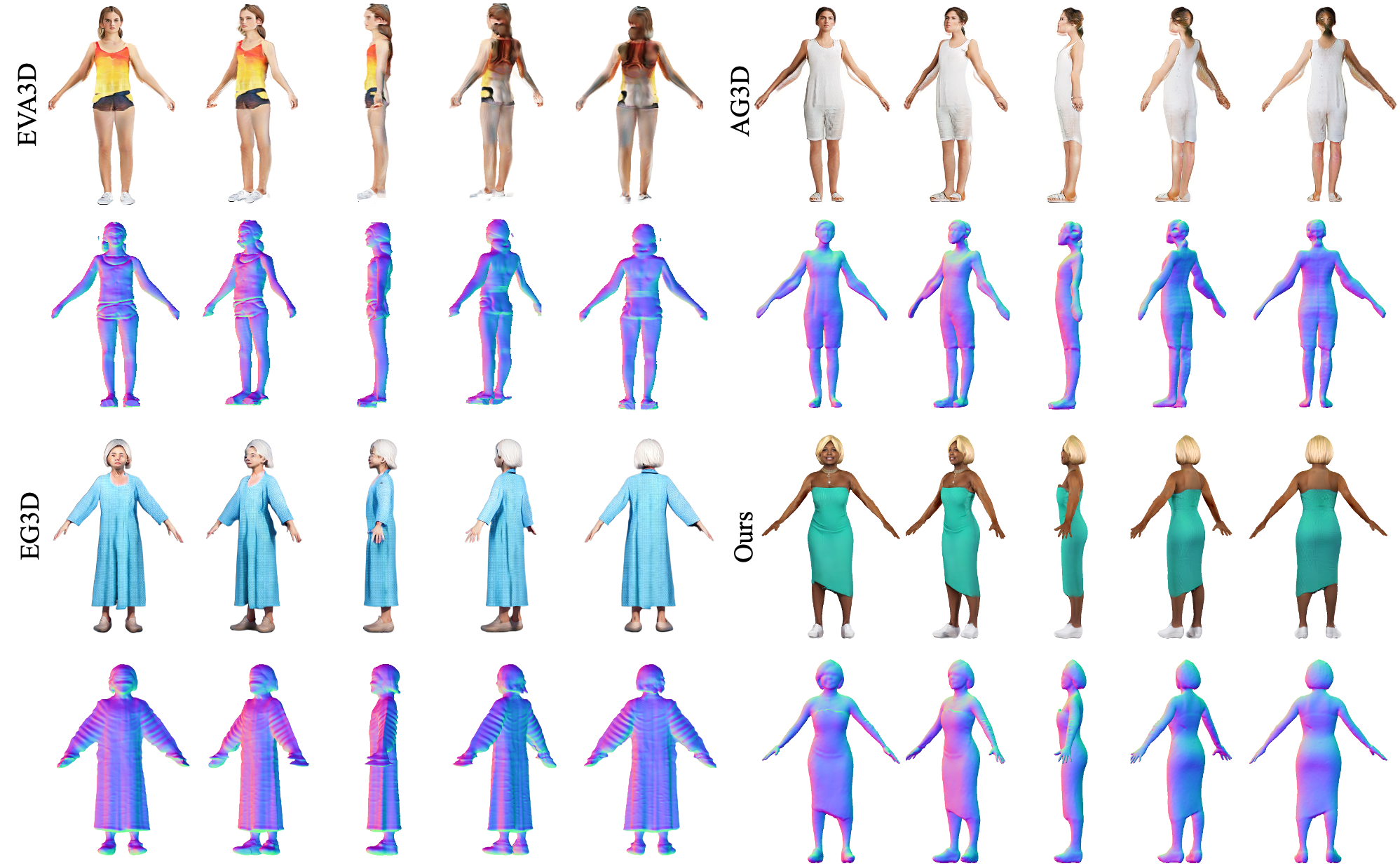}
\caption{Qualitative comparison with three state-of-the-art methods: EVA3D~\cite{hong2022eva3d}, AG3D~\cite{dong2023ag3d} and EG3D~\cite{chan2022efficient}.}
\label{fig:compare}
\end{center}
\end{figure*}

\subsection{Comparisons}
\noindent {\bf Qualitative comparison.} 
In Figure~\ref{fig:compare}, we compare our method with three baselines: EVA3D~\cite{hong2022eva3d} and AG3D~\cite{dong2023ag3d}, which are state-of-the-art methods for generating 3D humans from 2D images, and EG3D~\cite{chan2022efficient}, which serves as the foundational backbone of our method. 
The results of first two methods are produced by directly using source codes and trained models released by authors. We train EG3D using our synthetic images with estimated cameras from scratch. As we can see, EVA3D fails to produce $360^{\circ}$ humans with reasonable back inferring. AG3D and EG3D are able to generate $360^{\circ}$ renderings but both struggle with photorealism and capturing detailed shapes. 
Our method synthesizes not only higher-quality, view-consistent $360^{\circ}$ images but also higher-fidelity 3D geometry with intricate details, such as irregular dresses and haircuts.

\begin{table}
\setlength{\abovecaptionskip}{0.1cm}
  \centering
    \caption{Quantitative evaluation using FID, IS-360, normal accuracy (Normal) and identity consistency (ID). 
    }
    \small
  \begin{tabular}{ccccc}
    \toprule
    Method &  FID $\downarrow$  &  IS-360 $\uparrow$  & Normal $\downarrow$ & ID$\uparrow$  \\
    \midrule
    EVA3D~\cite{hong2022eva3d} & 15.91 & 3.19 & 30.81  & 0.72 \\
     AG3D~\cite{dong2023ag3d} & 10.93 & 3.28  & 20.83  & 0.69 \\
    EG3D~\cite{chan2022efficient} & 7.48  & 3.26 & 12.74  & 0.71 \\
    Ours & \textbf{2.73}  & \textbf{3.43} & \textbf{5.62} & \textbf{0.74}  \\
    \bottomrule
  \end{tabular}
  \label{table:quantitative}
\end{table}

\noindent {\bf Quantitative comparison.} 
Table~\ref{table:quantitative} provides quantitative results comparing our method against the baselines. We measure image quality with Frechet Inception Distance (FID)~\cite{heusel2017gans} and Inception Score~\cite{salimans2016improved} for $360^{\circ}$ views (IS-360). 
FID measures the visual similarity and distribution discrepancy between 50k generated images and all real images. IS-360 focuses on the self-realism of generated images in $360^{\circ}$ views. 
For shape evaluation, we compute FID between rendered normals and pseudo-GT normal maps (Normal), following AG3D. The FID and Normal scores of EVA3D and AG3D are directly fetched from their reports.
Additionally, we access the multi-view facial identity consistency using the ID metric introduced by EG3D. 
Our method demonstrates significant improvements in FID and Normal, bringing the generative human model to a new level of realistic $360^{\circ}$ renderings with delicate geometry while also maintaining state-of-the-art view consistency.

\subsection{Ablation study}

\noindent {\bf Synthesis flow and patch-composed rendering.} 
We assess the impact of our carefully designed synthesis flow by training a model with synthetic images but with camera and pose parameters estimated by SMPLify-X~\cite{pavlakos2019expressive} (w/o SYN-P). As Table~\ref{table:ab} shows,  the model w/o SYN-P results in worse FID and IS-360 scores, indicating that the synthesis flow contributes to more accurate physical parameters for realistic appearance modeling. By utilizing patch-composed rendering (PCR), the networks focus more on the human region, leading to more realistic results.

\begin{table}
\setlength{\abovecaptionskip}{0.1cm}
  \centering
    \caption{
    Results of models trained by replacing physical parameters with estimated ones  (w/o SYN-P) or removing patch-composed rendering (w/o PCR).
    }
    \small
  \begin{tabular}{cccc}
    \toprule
       & Ours  &  Ours-w/o SYN-P  & Ours-w/o PCR  \\
    \midrule
      FID $\downarrow$ & \textbf{2.73} & 4.28 & 3.26   \\
      IS-360 $\uparrow$  & \textbf{3.43} & 3.31  & 3.35   \\
    \bottomrule
  \end{tabular}
  \label{table:ab}
\end{table}

\begin{figure}
\begin{center}
\setlength{\abovecaptionskip}{0cm}
\setlength{\belowcaptionskip}{-0.3cm}
\includegraphics[width=1\linewidth]{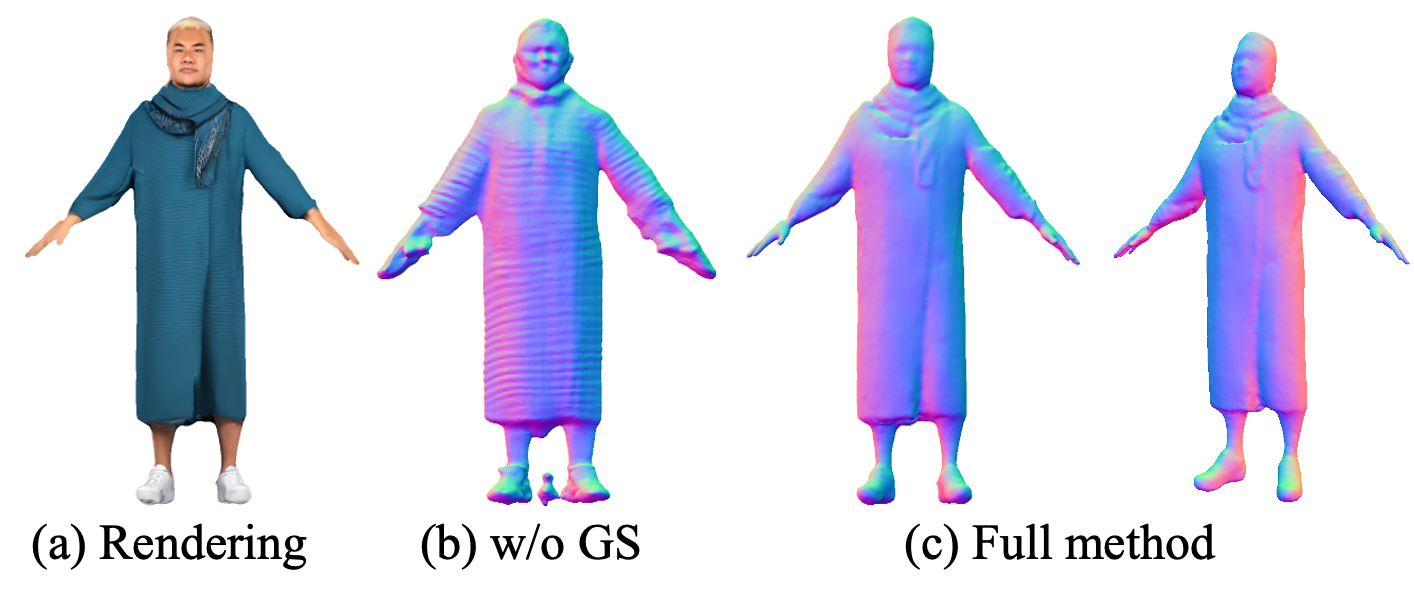}
\caption{Effects of the GS module to carve fine-grained surfaces. }
\label{fig:ab_geo}
\end{center}
\end{figure}

\noindent {\bf Geometry sculpting module (GS).}
We demonstrate the importance of this module by visualizing the meshes before and after its implementation. 
Figure~\ref{fig:ab_geo} (b) shows that the preceding module yields a coarse mesh due to the complex human anatomy and the challenges posed by decomposing ambiguous 3D shapes from 2D images. 
The GS module utilizes high-quality multi-view outputs and employs a more flexible hybrid representation to create expressive humans with arbitrary topologies. 
It learns from pixel-level surface supervision, leading to a significant improvement in shape quality, characterized by smooth surfaces and intricate outfits (Figure~\ref{fig:ab_geo} (c)).

\begin{figure}
\begin{center}
\setlength{\abovecaptionskip}{0cm}
\setlength{\belowcaptionskip}{-0.3cm}
\includegraphics[width=1\linewidth]{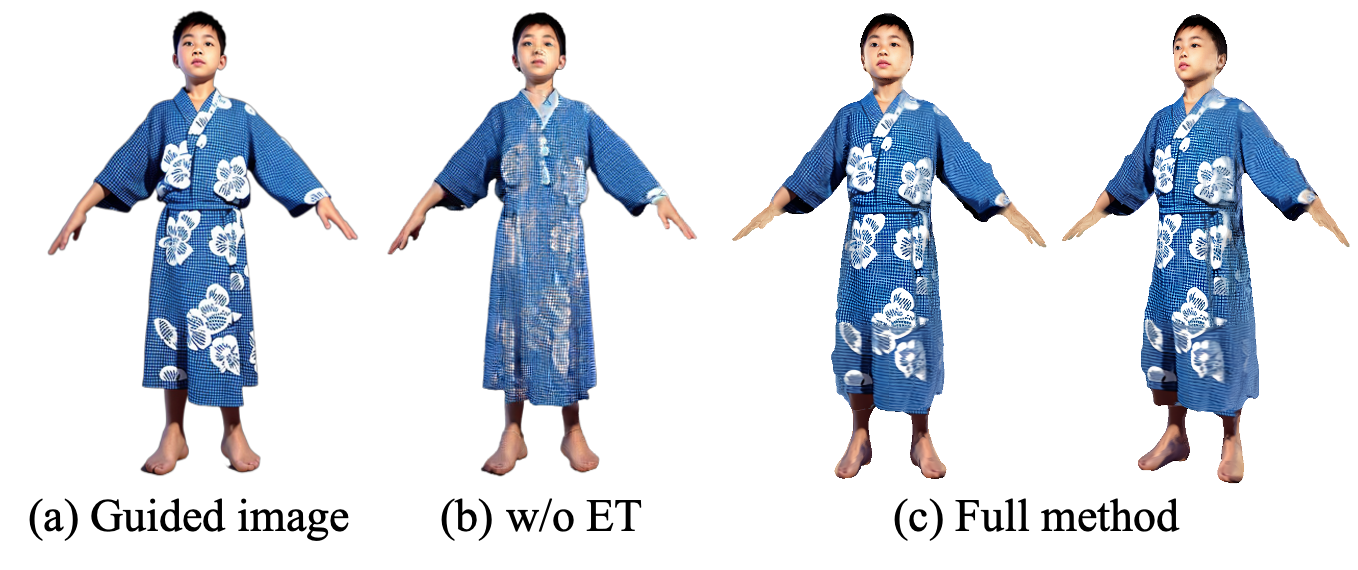}
\caption{Effects of the ET module for guided synthesis.}
\label{fig:ab_tex}
\end{center}
\end{figure}

\noindent {\bf Explicit texturing module (ET).}
This intuitive module not only extracts the explicit UV texture for complete 3D assets but also enables high-fidelity results for image guided synthesis. Following the flowchart in Figure~\ref{fig:infer}, we compare the results produced with and without this module. Our method without ET directly generates implicit renderings through PTI inversion, as shown in Figure~\ref{fig:ab_tex} (b). While it successfully preserves global identity, it struggles to synthesize highly faithful local textures (e.g., floral patterns). The ET module offers a convenient and efficient way to directly interact with the 3D representation, enabling the production of high-fidelity 3D humans with more consistent content including exquisite local patterns (Figure~\ref{fig:ab_tex} (a, c)).

\subsection{Applications}

\label{sec:application}

\noindent {\bf Avatar animation.}
All avatars produced by our method are in a canonical body pose and aligned to an underlying 3D skeleton extracted from SMPL-X. This alignment allows for easy animation and the generation of motion videos, as demonstrated in Figure~\ref{fig:teaser} and Suppl.

\noindent {\bf Texture doodle and local editing.}
Our approach benefits from explicitly disentangled geometry and texture, enabling flexible editing capabilities. Following the flowchart of text or image guided synthesis (Section~\ref{sec:infer}), users can paint any pattern or add text to a guided image. These modifications can be transferred to 3D human models by inputting modified views into the texture module (e.g., painting the text 'hey' on a jacket as shown in Figure~\ref{fig:teaser} (d)). Our approach also allows for clothing editing by simultaneously injecting edited guide images with desired clothing into the GS and ET modules (e.g., changing a jacket and jeans to bodysuits in Figure~\ref{fig:teaser} (e)). More results can be found in Suppl. 

\begin{figure}
\begin{center}
\setlength{\abovecaptionskip}{0.1cm}
\setlength{\belowcaptionskip}{-0.2cm}
\includegraphics[width=1\linewidth]{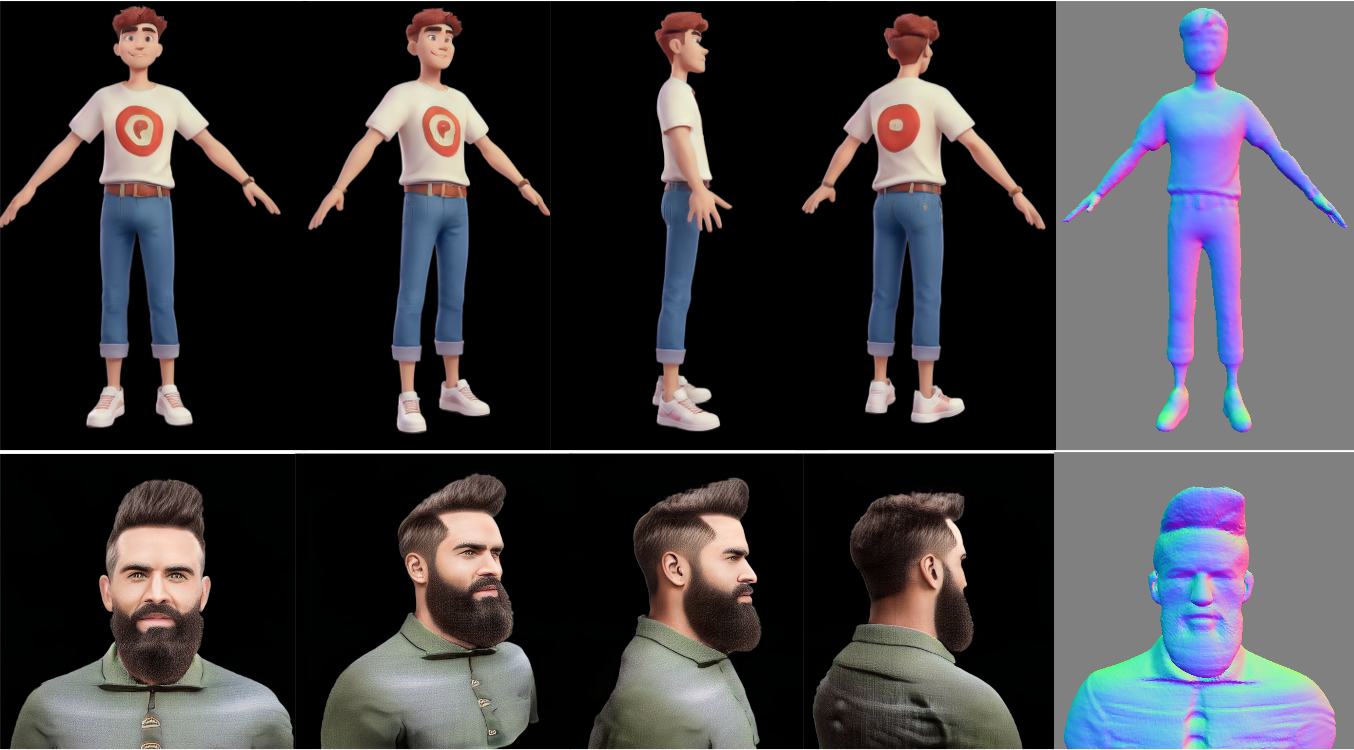}
\caption{Results synthesized by adapting our method to various styles (e.g., Disney cartoon characters) or contents (e.g., portrait heads).}
\label{fig:extension}
\end{center}
\end{figure}

\noindent {\bf Content-style free adaption. }
Our proposed scheme is versatile and can be extended to generate various types of contents (e.g., portrait heads ) and styles (e.g., Disney cartoon characters). To achieve this, we fine-tune our model using synthetic images from these domains, allowing for flexible adaptation. We showcase the results in Figure~\ref{fig:extension}. More results and other discussions (e.g., limitations, negative impact, etc.) can be found in Suppl.

\section{Conclusions}
We introduced En3D, a novel generative scheme for sculpting 3D humans from 2D synthetic data.
This method overcomes limitations in existing 3D or 2D collections and significantly enhances the image quality, geometry accuracy, and content diversity of 
generative 3D humans.
En3D comprises a 3D generative module that learns generalizable 3D humans from synthetic 2D data with accurate physical modeling, and two optimization modules to carve intricate shape details and disentangle explicit UV textures with high fidelity, respectively. 
Experimental results validated the superiority and effectiveness of our method. We also demonstated the flexibility of our generated avatars for animation and editing, as well as the scalability of our approach for synthesizing portraits and Disney characters. We believe that our solution could provide invaluable human assets for the 3D vision community. 
Furthermore, it holds potential for use in common 3D object synthesis tasks.

\section*{Acknowledgements}
We would like to thank Mengyang Feng and Jinlin Liu for their technical support on guided 2D image synthesis.

{
    \small
    \bibliographystyle{ieeenat_fullname}
    \bibliography{main}
}


\end{document}

%% file: preamble.tex
\usepackage[dvipsnames]{xcolor}
